\documentclass[10pt,a4paper,twoside]{tau-book}
\usepackage[english]{babel}
\usepackage{tau}
\usepackage{multicol}
\usepackage{enumitem}
\usepackage{float}
\usepackage{graphicx}

\title{BOSC: A toolbox for aerial imagery mapping}

\author[1]{Ricard Durall}
\author[ ]{Laura Montilla}
\author[ ]{Esteban Durall}

\affil[1]{ricard.durall.lopez@gmail.com}
\professor{}

\theabstract{
Accurate and efficient label of aerial images is essential for informed decision-making and resource allocation, whether in identifying crop types or delineating land-use patterns. The development of a comprehensive toolbox for manipulating and annotating aerial imagery represents a significant leap forward in remote sensing and spatial analysis. 
In this report, we introduce BOSC, a toolbox that enables researchers and practitioners to extract actionable insights with unprecedented accuracy and efficiency, addressing a critical need in today's abundance of drone and satellite resources.

For more information or to explore BOSC, please visit our \href{https://github.com/RicardDurall/BOSC_toolbox}{repository}.
}


\begin{document}
	
    \maketitle
    \abscontent
    \thispagestyle{firststyle}


\section{Introduction}
In recent years, advancements in technology have made aerial imagery indispensable across diverse domains, spanning agriculture \cite{onishi2021explainable}, urban planning \cite{muhmad2023uav}, and environmental monitoring \cite{son2021applications}. The accessibility and cost-effectiveness of acquiring high-resolution aerial images have surged, thanks to the proliferation of resources like drones and satellites. However, harnessing the full potential of this visual data necessitates sophisticated tools capable of efficiently processing and labeling images to derive meaningful insights.

To address the increasing demand for streamlined aerial image analysis, we introduce BOSC, a comprehensive toolbox tailored to the evolving needs of researchers and practitioners. BOSC offers a versatile range of functionalities designed specifically for manipulating and labeling aerial images with exceptional precision and ease. Notably, it enables the drawing of masks and the assignment of class labels on aerial images, whether through manual or automatic means, leveraging cutting-edge artificial intelligence models.

\section{BOSC Toolbox}
\subsection{Structure}
\begin{figure}[t]
    \centering
    \includegraphics[width=0.9\columnwidth]{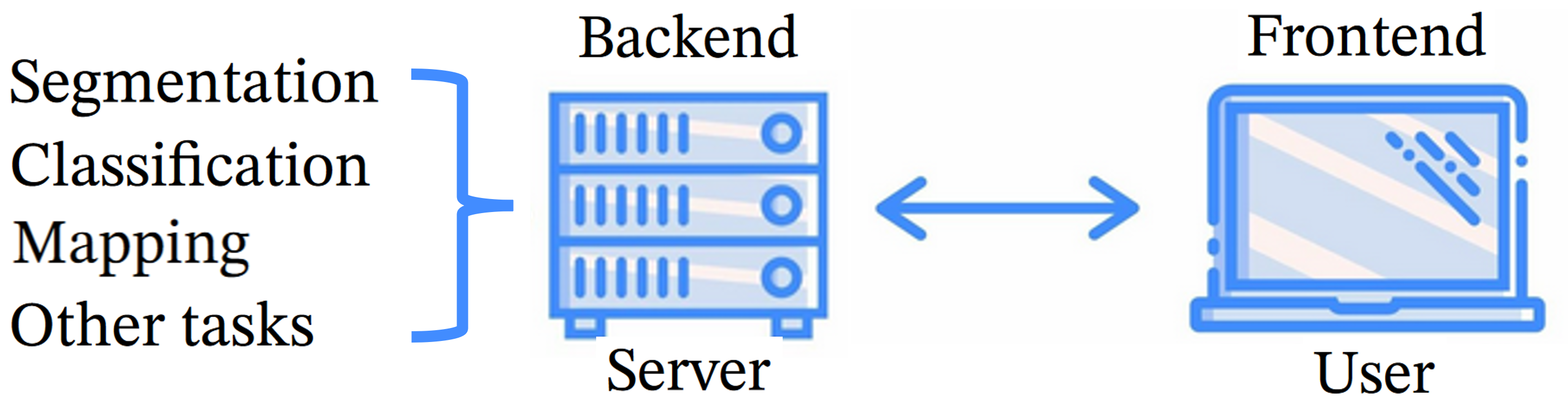}
    \caption{Client-Server architecture of BOSC.}
    \label{fig:struc}
\end{figure}

This toolbox consists of two primary components: the backend and the frontend (see Figure \ref{fig:struc}).

\textbf{Backend:} At the heart of the toolbox lies the backend, where all heavy computation operations are executed in a server. This includes tasks such as segmentation, clustering, mapping, and other complex analytical processes. By centralizing these operations in the backend, we ensure swift and accurate processing of large datasets without compromising on performance. Leveraging powerful computational resources, the backend handles the computational heavy lifting, allowing users to interact with the toolbox seamlessly.

\textbf{Frontend:} Complementing the backend, the frontend serves as the user interface, providing an intuitive platform for users to access and utilize the toolbox's functionalities. Here, users can easily upload and manipulate aerial images, perform annotations, and visualize analysis results in real-time. The frontend is designed with simplicity and ease of use in mind, shielding users from the technical complexities of the underlying computational processes. By offering a streamlined and user-friendly interface, the frontend enhances the overall user experience, enabling users to focus on their tasks without being encumbered by unnecessary complexities.

\subsection{Method}
This toolkit comprises three primary methods.

\textbf{Segmentation Engine:}
Our implementation presents a robust and generalizable image segmentation model utilizing advanced deep learning techniques. Drawing inspiration from the Segment Anything model \cite{kirillov2023segment} and Fast Segment Anything \cite{zhao2023fast} model, our approach effectively captures intricate spatial dependencies and hierarchical features within images. This ensures high accuracy in segmenting a diverse array of objects. Trained on extensive labeled datasets, the model delivers pixel-wise predictions during inference, enabling rigorous analysis and supporting various downstream applications like object tracking and scene comprehension. Engineered for efficiency and scalability specifically for aerial imagery, our model consistently integrates into existing workflows, offering users potent tools for thorough image analysis in aerial domains.

\begin{figure*}[t]
    \includegraphics[width=\textwidth]{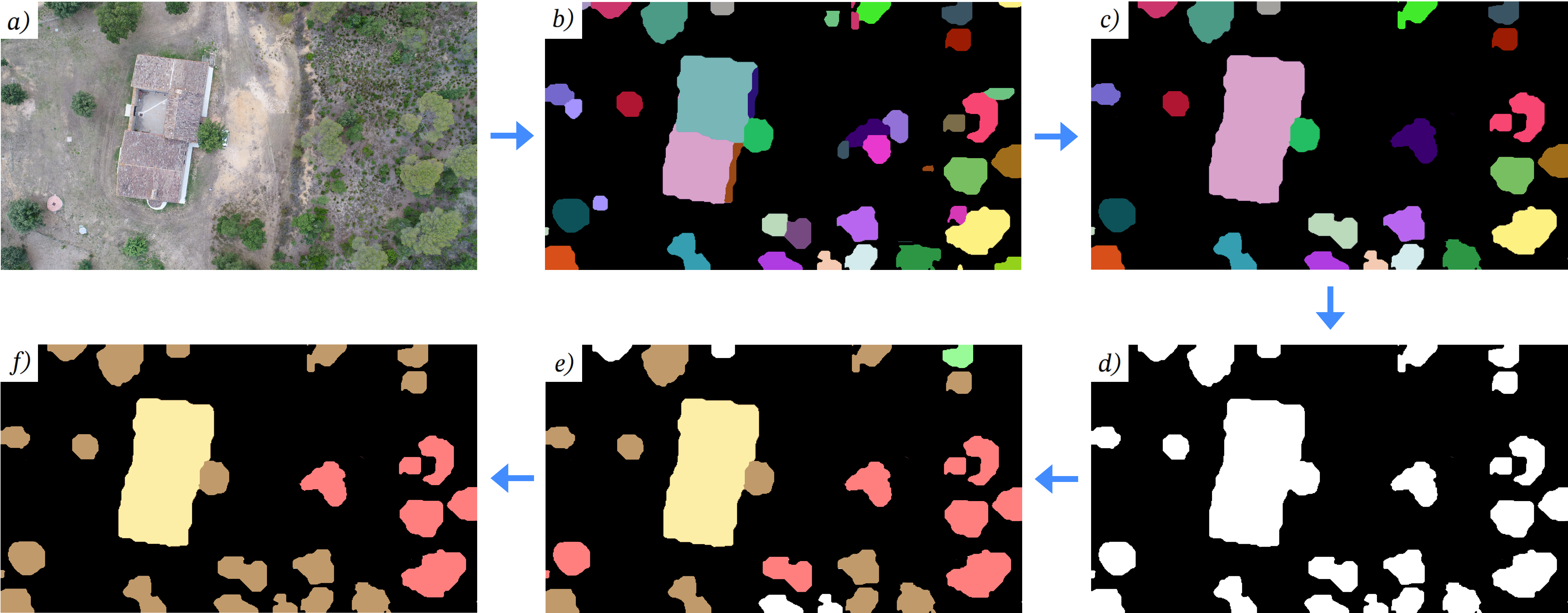}
    \caption{Example of crown-tree classification: a) Raw input image. b) Automatic segmentation output. c) Manually fine-tuned segmentation output. d) Default classification. e) Automatic classification output (no labeled object provided). f) Manually fine-tuned classification output.}
    \label{fig:pipeline}
\end{figure*}

\begin{figure*}[t]
    \includegraphics[width=\textwidth]{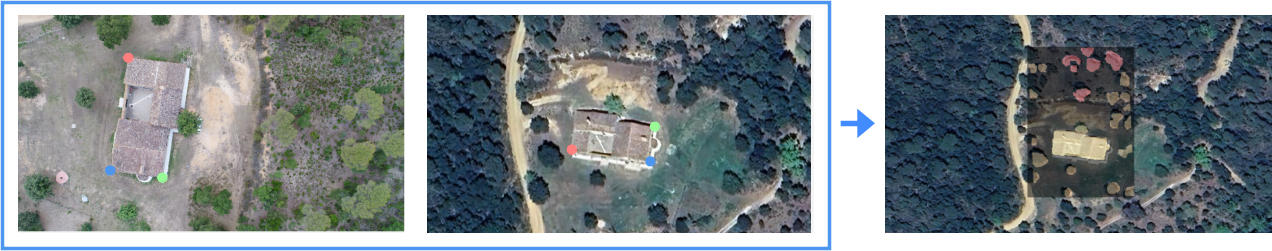}
    \caption{Example of mapping between an input image (left) and its corresponding location on a third-party map (center). Both images display three selected points used for the mapping. The final outcome (right) shows the new layer containing the labeld segmentation image overlaid on the interactive map.}
    \label{fig:mapping}
\end{figure*}

\textbf{Classification Engine:}
Our classification engine utilizes unsupervised classification techniques, specifically a hirearchical clustering algorithm \cite{zepeda2013hierarchical}, for object classification. Operating without the need for labeled data, it discerns patterns by extracting features from pre-trained ConvNeXt model \cite{liu2022convnet} and grouping similar features into clusters. Additionally, users have the option to provide labeled objects to guide and refine classification accuracy. The user-friendly interface enables interactive input definition and visualization of clustering results, regardless of the complexity of the data. This clustering approach presents a robust solution for object identification and classification in image analysis tasks.

\textbf{Mapping Engine:}
Powered by \cite{opencv_library}, our mapping function employs geometric transformations to align the user input image with a reference map. To achieve that users select identifiable features (3 points) common to both the image and the map, serving as control points for transformation. Affine transformations are applied to warp the image, adjusting for translation, rotation, scaling, and shearing. Additional refinement steps ensure accurate alignment, enhancing spatial analysis reliability. The implementation prioritizes efficiency and accuracy, utilizing optimized algorithms for rapid and precise transformations. The user interface allows intuitive point selection and real-time visualization of mapped results. 

\section{Pipeline}

To demonstrate the functionality of BOSC, we present a practical example of crown-tree classification, outlining the process from input image to mapping the labeled segmentation solution. In Figure \ref{fig:pipeline}, starting from the upper left corner and moving clockwise, we have:

\begin{enumerate}[label=\textbf{\alph*)}, itemsep=0pt,parsep=2pt]
    \item \textbf{Selection:} The raw input image containing the objects of interest, such as an aerial image capturing a landscape, agricultural field, or burban area.
    \item \textbf{Segmentation:} The chosen input is processed by a segmentation engine, which identifies individual objects within the image. This results in a segmentation image, where each color corresponds to a separate object.
    \item \textbf{Refined Segmentation:} Users have the flexibility to correct any segmentation errors by utilizing the drawing palette as needed.
    \item \textbf{Initial Classification:} By default, all segmented objects are initially assigned to the same class (label), visually represented by white color.
    \item \textbf{Classification:} After the segmentation step, the classification algorithm generates an initial solution, assigning a color (representing a class) to each segmented object. For example, yellow for buildings and brown for oaks.
    \item \textbf{Refined Classification:} Once again, users have the flexibility to adjust any incorrect classes to produce the final labeled segmentation image.
\end{enumerate}

After the completion of the classification process, the next step employs the mapping egine, as depicted in Figure \ref{fig:mapping}.
\begin{enumerate}[itemsep=0pt,parsep=2pt]
    \item[] \textbf{Mapping:} Our toolbox enables the seamless overlay of our labeled segmentation image onto a third-party map. This is achieved by selecting three common points in both the raw input image and the target map.
    \item[] \textbf{Final Result:} Finally, we can visualize our results on the map, providing a comprehensive view of the data. This includes relevant statistical information, such as the number of instances of a particular class and the total coverage area of specific classes, and it can be easily extended to other pertinent metrics. This detailed visualization allows for a deeper understanding and analysis of the mapped data.
\end{enumerate}

\begin{figure*}[htbp]
    \centering
    \includegraphics[width=\textwidth]{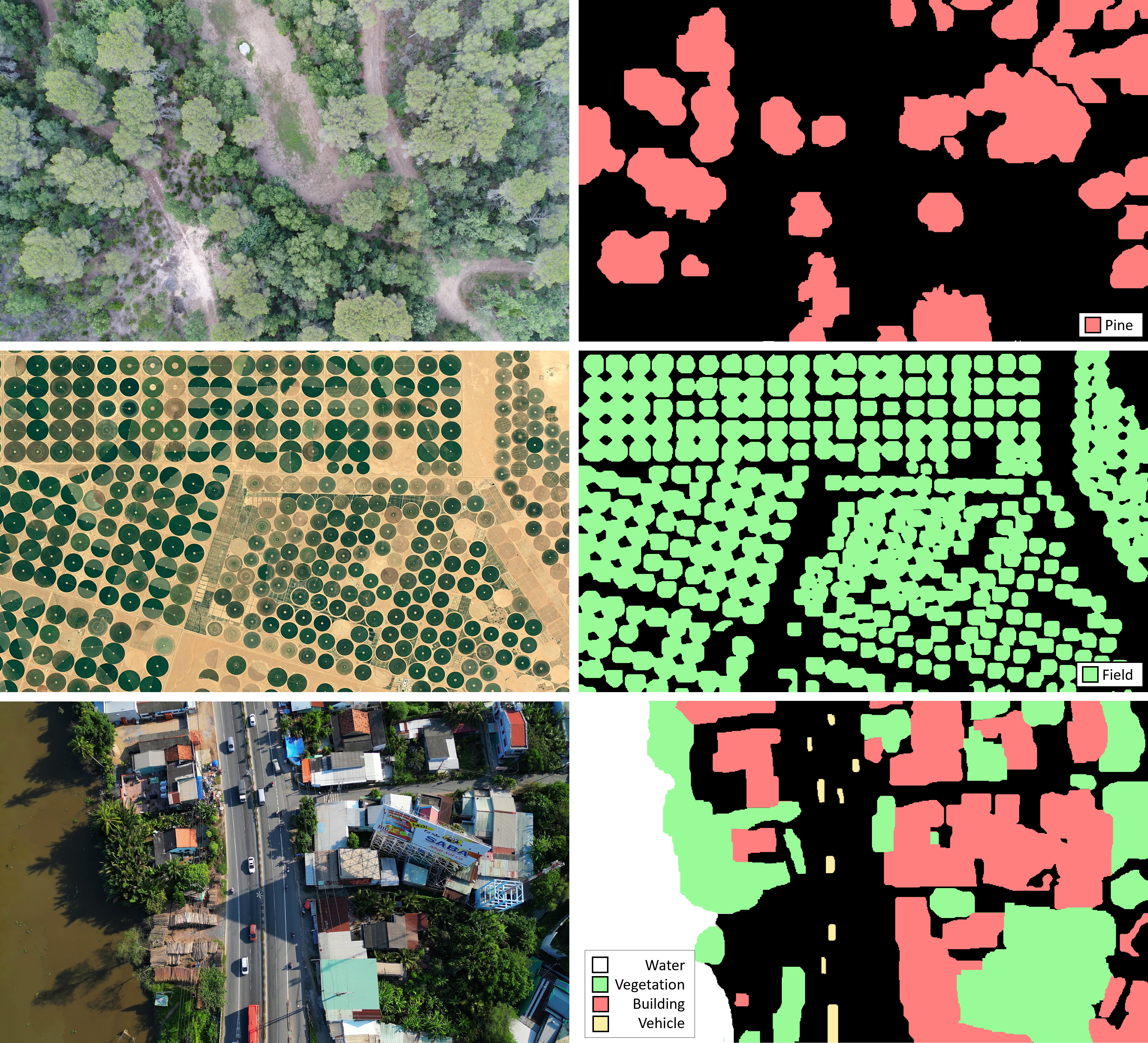}
    \caption{From top to bottom, the first pair of images illustrates an example where BOSC is employed to map only pine trees. This scenario is useful for forest inventory, particularly in estimating the carbon fixation coefficient (net oxygen production), which is tree-dependent. The second example depicts irrigated agriculture in the desert (1 km diameter) in Saudi Arabia. By using the segmentation mask, the status of the crops can be monitored and farming practices optimized. Finally, the third example demonstrates a scenario where multiple classes are mapped, showcasing BOSC's capability to handle diverse scenarios such as semi-urban or urban environments.}
    \label{fig:example}
\end{figure*}

\section{Results}
In this section, we present various outcomes generated using BOSC. As illustrated in Figure \ref{fig
}, this toolbox is versatile and adaptable to multiple scenarios, as its internal engines are designed to generalize independently of the target domain. This adaptability makes BOSC highly flexible, enabling its efficient use in agricultural settings for tasks such as forest population management and optimization of agricultural plantations. Additionally, we explore further use cases, such as the classification of urban or semi-urban images, highlighting the tool's extensive applicability and versatility.

\section{Conclusions}
In summary, the BOSC toolbox offers a comprehensive solution for aerial image analysis, incorporating advanced segmentation, classification, and mapping functionalities. By streamlining the image processing workflow and providing intuitive user interfaces, BOSC enables efficient and accurate classification of objects within images. Additionally, its ability to incorporate user inputs and refine segmentation results enhances flexibility and adaptability. Overall, BOSC represents a significant advancement in remote sensing and spatial analysis, empowering researchers and practitioners with powerful tools for extracting actionable insights from aerial imagery.

\printbibliography

\end{document}